%% file: root.tex
\documentclass[letterpaper, 10 pt, conference]{ieeeconf}  

\IEEEoverridecommandlockouts                              

\usepackage{amsmath}
\usepackage{dblfloatfix}
\usepackage{hyperref}
\usepackage{listings}
\usepackage{mathtools}
\usepackage{pgfplots}
\usepackage{pifont}
\usepackage{siunitx}
\usepackage{subcaption}
\usepackage{tikz}
\usepackage{tikzscale}
\usepgfplotslibrary[groupplots,dateplot]
\usetikzlibrary{patterns,shapes,arrows}
\pgfplotsset{compat=newest}

\usepackage{atbegshi}

\newlength\figureheight
\newlength\figurewidth

\newcommand*{\qmlrosplugin}{\textit{qml\_ros\_plugin}}

\newcommand*{\hectorrvizoverlay}{\textit{hector\_rviz\_overlay}}
\newcommand*{\rviz}{\emph{RViz}}

\newcommand*{\new}[1]{\textcolor{red}{#1}}
\renewcommand*{\new}[1]{#1}

\usepackage[sortcites,backend=biber,hyperref=true]{biblatex}
\bibliography{bibliography}

\title{\vspace{-2cm}%
\small Preprint of the paper which appeared in:\\ \large IEEE International Symposium on Safety, Security, and Rescue Robotics (SSRR) 2023%
\vspace{1cm}\newline\LARGE\bf%
Hector UI: A Flexible Human-Robot User Interface for (Semi-)Autonomous Rescue and Inspection Robots
}

\author{Stefan Fabian and Oskar von Stryk
  \thanks{Stefan Fabian and Oskar von Stryk are with the Technical Uni\-versity of Darmstadt,
    Computer Science Department, Simulation, Systems Optimization and Robotics Group, Germany.\newline
    {\tt\small \{fabian,stryk\}@sim.tu-darmstadt.de}}%
  \thanks{%
    Research presented in this paper has been supported in parts by the LOEWE initiative (Hesse, Germany) within the emergenCITY center and by the German Federal Ministry of Education and Research (BMBF) within the DRZ project (grant no. 13N16475) and the KIARA project (grant no. 13N16274).%
    \newline 979-8-3503-8111-5/23/\$31.00 \textcopyright 2023 IEEE%
    \newline DOI: 10.1109/SSRR59696.2023.10499954%
  }%
}

\begin{document}

\relax\maketitle
\thispagestyle{empty}
\pagestyle{empty}

\begin{abstract}
The remote human operator's user interface (UI) is an important link to make the robot an efficient extension of the operator's perception and action.
In rescue applications, several studies have investigated the design of operator interfaces based on observations during major robotics competitions or field deployments.
Based on this research, guidelines for good interface design were empirically identified.
The investigations on the UIs of teams participating in competitions are often based on external observations during UI application, which may miss some relevant requirements for UI flexibility.
In this work, we present an open-source and flexibly configurable user interface based on established guidelines and its exemplary use for wheeled, tracked, and walking robots.
We explain the design decisions and cover the insights we have gained during its highly successful applications in multiple robotics competitions and evaluations.
The presented UI can also be adapted for other robots with little effort and is available as open source.
\end{abstract}

\section{INTRODUCTION}
\input{sections/introduction.tex}

\section{DESIGN}
\label{sec:design}

\input{sections/design.tex}

\section{CONTROL}
\label{sec:control}
\input{sections/control.tex}

\section{IMPLEMENTATION}
\label{sec:implementation}
\input{sections/configuration.tex}
\section{APPLICATIONS}
\label{sec:applications}
\input{sections/applications.tex}

\addtolength{\textheight}{-2cm}   

\section{LESSONS LEARNED}
\label{sec:insights}
All of the above challenges had a lot of shared requirements for the UI but depending on the task the importance of different visualizations and functionalities varied.

For mobile navigation, it is of high importance to have low-latency information such as camera streams and 3D information can be helpful in narrow spaces but with good camera placement, it is not always necessary.
For example, the Asterix robot has cameras that point slightly downward so the flippers are in view and can be used to judge distances near the robot in narrow spaces, e.g., if a doorway is wide enough for the robot to pass through.
Additional depth information in the cameras will also have a greater effect than a separate 3D view in time-critical situations as the operator may not be able to focus on multiple areas.
In our experience, driving solely based on the 3D view is only possible if the environment representation is reasonably accurate and both the map and the robot state are provided at low latency.
While driving with the eC Scout robot through the mostly flat nuclear plant, the 3D view proved more useful than the cameras for judging if a doorway was wide enough to fit through as the ground geometry was not relevant and a 2D map combined with a cropped pointcloud as shown in Fig.~\ref{fig:scout_enrich} was sufficient to obtain all the information necessary for navigation.
In the RoboCup Rescue League, on the contrary, the terrain is very complex and the navigation and robot pose are often unstable.
Therefore, the robot state has to be very accurate and the environment representation has to accurately capture the ground structure in a way that allows identifying viable paths.
Unlike with camera views it can be hard to judge whether the expected state of the robot and the real state match resulting in low trust in the 3D view for navigation in time-critical situations.

Manipulation in confined spaces in contrast requires accurate depth information that is easier to obtain from a 3D view than camera views.
During door opening, we used the raw pointcloud to judge how close the robot and gripper are to the door while moving the base before using the gripper camera to position the gripper at the handle.
Additionally, in this task, we profited strongly from generic simple robot actions such as moving the gripper in a door opening pose which in conjunction with the 3D view enabled us to perform best in the door opening task in 2023 despite having the robot with the shortest arm barely able to reach the door handle.

In manipulation tasks, we experienced mode error mistakes on multiple occasions because the operator thought the gamepad control was in manipulation when in reality it was in driving mode which resulted in minor collisions with the environment.
To address this, we have added an extra manipulation operation mode with a different color scheme to make the current mode more visible.

\section{CONCLUSIONS}
In this paper, we have presented a flexible UI that can be used with different types of rescue and inspection robots and their functionalities with low or manageable effort for any needed adaptions.
It has been successfully applied and evaluated for different types of robots in four different types of international robotics competitions.
As the UI is available as open source, it can be applied and adapted to further types of robots.
The code for the user interface can be found at \url{https://github.com/tu-darmstadt-ros-pkg/hector_user_interface}.










\renewcommand*{\bibfont}{\small}
\printbibliography

\end{document}

%% file: sections/introduction.tex
The response to natural and man-made disasters can be extremely dangerous for the human rescue forces but obtaining information about the current situation at hand is critical for planning and execution of any rescue or disaster response mission.
In many cases, an area or a building structure can not be inspected by first responders before risks are reduced, e.g.,  fires are under control or a building structure is reinforced.
A process that would take precious hours delaying activities for rescue or mitigation of hazards.
In these cases, robotics technology is increasingly being used to gather current information on sites where humans can not enter.
Since one of the first known uses of rescue robots at the World Trade Center disaster proved their value for rescue missions~\cite{murphy2004trialbyfire}, there has been increased research into functionalities that further improve the value of a robot in this application.
In rescue operations, the operational environments are very different, usually not known beforehand, and the tasks are diverse. Hence, many researchers are working on autonomous functionalities and assistance functions for specific tasks in these environments.
For testing and demonstration of new robot features, UIs are often created by developers who focus on the features developed rather than the needs of the end-users represented by human factors, general usability, and information visualization.

\begin{figure}[t]
  \vspace{2mm}
  \centering
  \includegraphics[width=\linewidth]{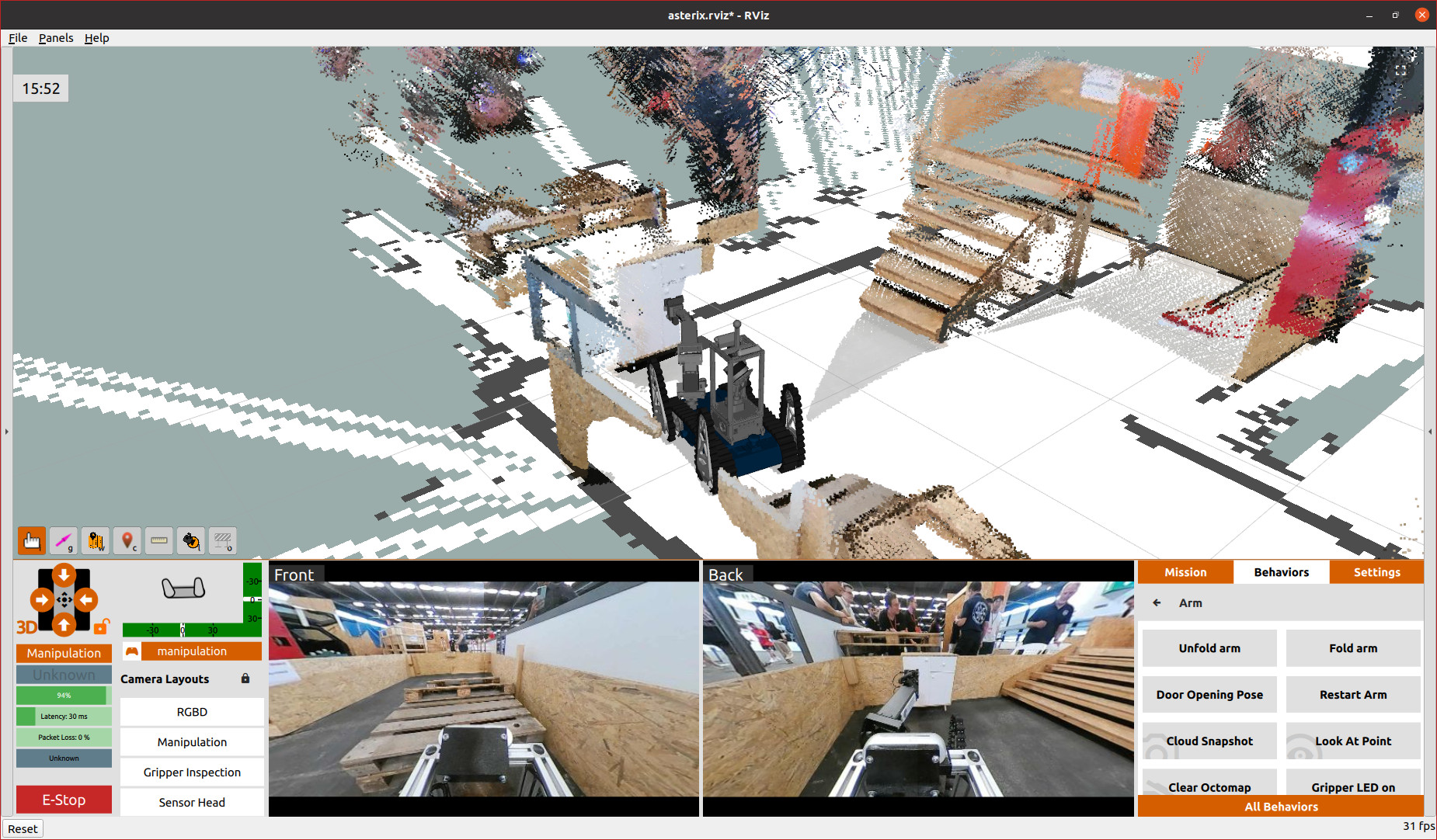}
  \caption{\new{The open-sourced interface presented in this paper in use at the international RoboCup Rescue 2023 competition.}}
  \label{fig:userinterface}
\end{figure}

As capable robotic platforms and sensors become more affordable, the use of robots with autonomous assistance functionalities is spreading to various industries and beginning to become economically viable for emergency responders.
As this transition is happening the remote operators of robotics systems will increasingly become end-users without a strong background in robotics.
Hence, there have been increasing calls for user-centered human-robot interfaces \cite{rea2022callforfocus}.

There are numerous studies on different interfaces employed by teams in competitions such as the DARPA Robotics Challenge \cite{norton2017darpa} and the RoboCup Rescue competition  \cite{yanco2007studyhrirescuecomp,scholtz2004evaluationhri}.
Additionally, we have field reports from actual deployments \cite{murphy2005hrilessonssar}.

Several of these papers observe the UIs of different teams from an outsider's perspective and empirically identify guidelines for good UI design based on the teams' performances.
Murphy and Tadokoro~\cite{murphytadokoro2019fieldrobotics} propose guidelines for good human-robot user interfaces in field robotics and a taxonomy categorizing human-robot interfaces into three groups: \textit{Developer}, \textit{End-User}, and \textit{Public}.
The \textit{End-User} group can be split further into the operator of the robot and mission specialists, e.g., structural engineers, medical specialists, etc.
Following the notion of~\cite{murphytadokoro2019fieldrobotics}, we will refer to the relevant design guideline as \textbf{Gi.k} with i, k being integers.

\begin{figure*}[b]
  \includegraphics[width=\linewidth]{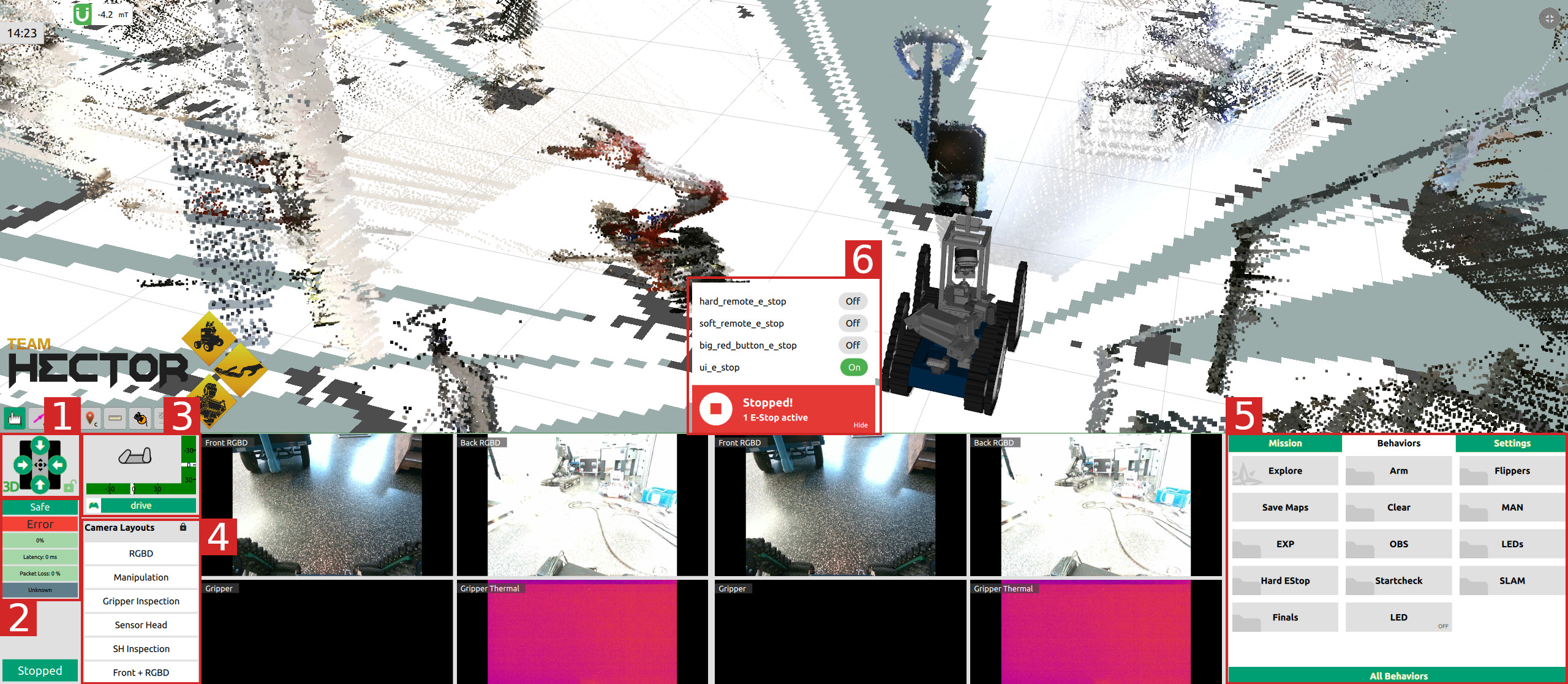}
  \caption{The operator interface for our robot Asterix while the UI E-Stop is active.}
  \label{fig:asterix}
\end{figure*}

In this paper, we present a flexible and versatile UI that meets different needs for a user-centered UI for non-expert robot operators as well as for developers.
In \ref{sec:design}, we explain the design of the components and the proposed UI framework, before briefly describing the control modes in \ref{sec:control}.
The implementation and configuration options to extend the UI with new capabilities or sensors, and adapt to different robot platforms are covered in \ref{sec:implementation}.
In \ref{sec:applications}, we briefly present the applications where the UI has been used and tested, as well as the respective competition-specific requirements.
Finally, in \ref{sec:insights}, we discuss lessons learned from applying the UI in various robotics competitions.

%% file: sections/design.tex
The user interface was developed as an overlay for the popular robotics data visualization tool \rviz~and is rendered using the open-source library \textit{hector\_rviz\_overlay} \cite{fabian2021hritools}.
Using this approach, it is possible to make use of the extensive library of community-developed visualization plugins.
Rather than developing separate applications for developers and end-users, we implement the end-user view as a special case of the developer view where we hide the advanced configuration options.
Following the consensus of HRI Guidelines~\cite{yanco2007studyhrirescuecomp,adamides2015usabilityguidelines,murphytadokoro2019fieldrobotics}, \textbf{G2.2} \textit{"The primary window should be largest, most captivating, and preferably
most central"} and \textbf{G2.3} \textit{"Avoid overlays or using pop up dialog boxes unless it is an absolutely
critical event"}, the basic interface is implemented as a single window application, however, it is possible to pop out camera views and move them to a second monitor when operating with multiple operators to increase the chance of finding victims or objects of importance~\cite{murphy2005hrilessonssar}.

In this section, we will discuss the design of the interface starting with a top-down overview before delving into the components and design decisions.

\subsection{Overview}
To avoid the issue of overlapping windows blocking information~\cite{yanco2007studyhrirescuecomp} and make efficient use of the available screen space, we render our user interface on top of the \rviz~3D view.
The interface is split into two main parts:

\subsubsection{Dashboard}
The dashboard contains robot status information, an emergency stop, robot state information, two camera views, and a control area to execute missions consisting of a sequence of tasks, trigger (semi-)autonomous behaviors, and change mission-relevant settings.

\subsubsection{3D View}
The \rviz~3D view is used to display the robot joint configuration and fused sensor data to give an understanding of the robot's environment including depth information that is hard to gather from the limited field-of-view of camera images~\cite{alfano1990fov}.
The joint configuration consists of the state of the four co-joined movable flippers to overcome obstacles and the manipulator arm.

\subsection{Dashboard}
Following \textbf{G2.4} \textit{"Group windows and widgets with spatial or semantic correspondence
together"}, we split the dashboard into three main areas of semantic correspondence.
From left to right: the robot status and state information, the camera views, and the control panel.

\subsubsection*{3D View Control}
In Fig. \ref{fig:asterix} (1), the control for the virtual camera of the 3D view is located.
The four buttons on the 2D robot illustration move the camera to the respective poses (left, front, right, back) relative to the current location of the virtual robot in the map frame with the focus lying in the robot base.
On the bottom left, a toggle allows switching between a 3D camera and an orthographic 2D top-down projection, and the lock symbol in the bottom right allows locking the camera wrt to the robot base.
When the camera is locked, it will move in the map frame keeping its position relative to the robot base.
It is still possible to change the orientation of the camera but moving the camera manually will automatically unlock.
To avoid motion sickness due to high-frequency motions, the locked camera will follow using a P controller to dampen the movements.

\subsubsection*{Robot Status}
The status of the robot is displayed in Fig.~\ref{fig:asterix} (2).
At the top, the current operation mode is shown which can be one of the following: Autonomous (blue), Teleoperation (orange), Manipulation (vermillion), or Safe (bluish-green).
The colors are visually distinct and optimized for color-blind individuals~\cite{wong2011color}.
Following \textbf{G3.3.3} \textit{"Maintain consistency (use of color, same fonts, same line type, etc.)
across windows and widgets"} and to reduce the risk of mode error~\cite{sarter1995modeerror}, the color scheme of the entire interface changes according to the current operation mode.
Below, a diagnostics summary with the aggregate states OK (green), WARNING (orange), and ERROR (red) is shown.
Clicking the summary will open a popup with the detailed status of the robot's components.
Lastly, the battery and connection state are displayed.
At the lower end, there is an emergency stop switch, which due to its limited reliability (as it uses ROS and standard network transmission) should only be used in non-critical situations, as triggering can not be guaranteed with sufficient certainty.

\subsubsection*{Robot State}
Obtaining and maintaining a mental model of the current flipper configuration while traversing challenging terrain is strenuous for the operator and therefore prone to errors. Hence, in Fig.~\ref{fig:asterix} (3), the current state of the robot's mobile base is displayed with a side-view of the flipper configuration and the robot's orientation (roll \& pitch).
Below, a mode indicator displays the current gamepad control mode to reduce mistakes due to mode error.
The gamepad control modes are explained in \ref{sec:gamepadcontrol}.

\subsubsection*{Camera Views}
Two camera views are used for live situation awareness.
To allow for extension and evolution of the robot hardware~\cite{yanco2004usability}, the camera overview is organized as a scrollable GridView and cameras can be added by advanced users or developers using a simple dialog.
During time-critical operations different views may be necessary and some tasks require switching between different views.
Following \textbf{G2.7} \textit{"Design for dynamic content if needed"} and \textbf{G4.2} \textit{"Minimize the number of clicks (or pull down menus) needed to reach a goal"}, both camera views can quickly be switched using a list of dynamically saved camera configurations left of the camera views as shown in Fig.~\ref{fig:asterix} (4).
If more than two cameras are required, a camera can also be opened in a separate window using the popout button.

During teleoperation or remote supervision, it is not only important to have a live video source of what is currently happening but also to know how up-to-date the information being displayed is.
Therefore, we display the received camera frame rate and latency directly in the corner of each camera view so that the operator can properly assess his or her situational awareness and reduce errors due to outdated information~\cite{endsley1988sa}.

\subsubsection*{Control Panel}
The rightmost part of the dashboard -- Fig.~\ref{fig:asterix} (5) -- is a tabview to control the robot's actions of varying complexity ranging from simple predefined joint configurations to autonomous exploration behaviors.
Three tabs allow control of the robot's (semi-)autonomous functions.

The first from left to right, is the \textit{Mission} tab which during an active mission displays the task list, highlighting the current task and a control area to change the execution state of the mission with one of the following actions: \textit{Go back to the previous task}, \textit{Pause / Resume}, \textit{Stop mission} (resets state), \textit{Skip current task}.

The second tab is the \textit{Actions} or \textit{Behaviors} tab and the default view on application-start.
At the bottom, a button opens the alphabetically sorted list of all available actions the robot may perform.
Actions are explained in more detail in \ref{sec:configuration}.
During operation not all of these actions are relevant to the current mission or task, hence, actions from this list can be dragged and dropped in the structure view above the button in the original view.

The structure view allows the free inserting, ordering, and grouping of the available actions in folders.
For example, manipulation-related actions -- such as unfolding the arm from its parking position, moving the arm to predefined joint configurations, or toggling the gripper LED -- can be put in a separate manipulation folder.
The same action can also be added at multiple locations in the hierarchical directory tree structure.

\begin{figure}[t]
  \vspace{2mm}
  \includegraphics[width=\linewidth]{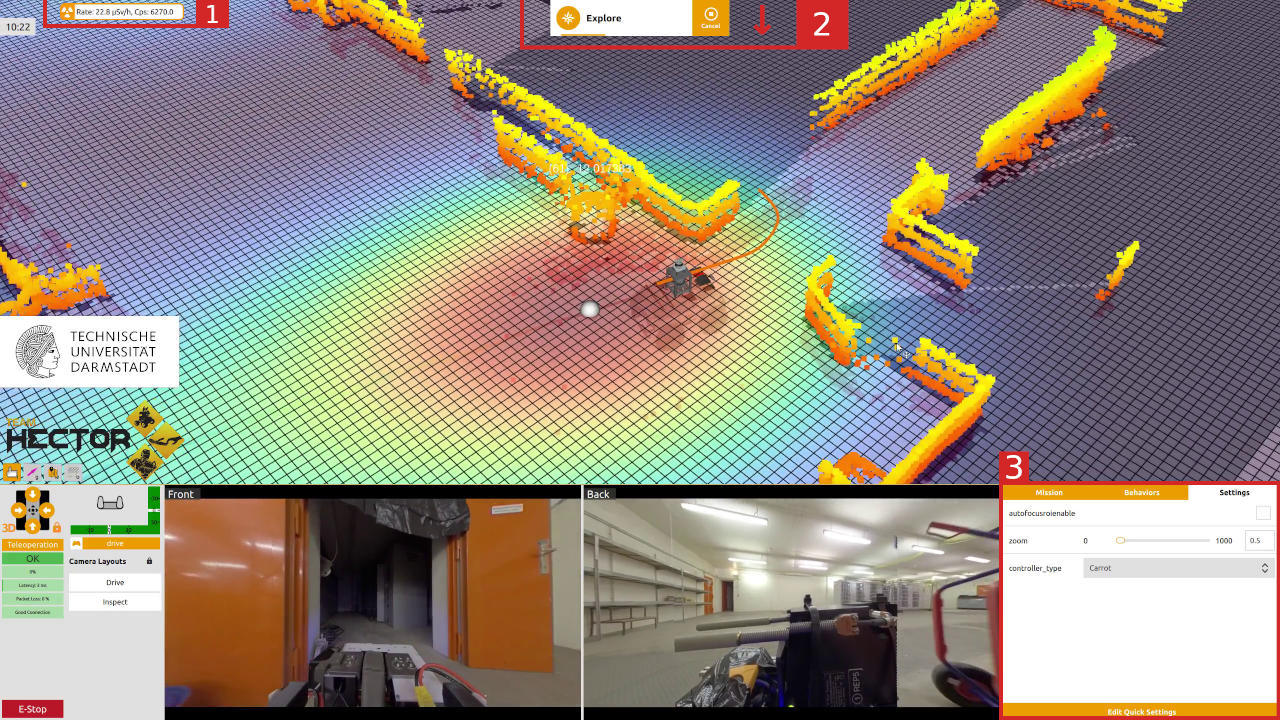}
  \caption{The eC Scout robot UI while exploring and mapping a part of a nuclear plant, and locating radioactive probes in the context of the ENRICH 2023 robotics hackathon.}
  \label{fig:scout_enrich}
\end{figure}

Finally, the right tab named \textit{Settings} contains selected parameters from \textit{Dynamic Reconfigure} that affect the execution of the (semi-)autonomous actions.
Settings can be added by advanced users and developers using a dialog with all available parameters and given a custom name to hide the implementation details from end-users.
The view of the settings tab for the end user is shown in Fig.~\ref{fig:scout_enrich}~(3).

\begin{figure*}[b!]
  \begin{center}
  \includegraphics[width=0.4\linewidth]{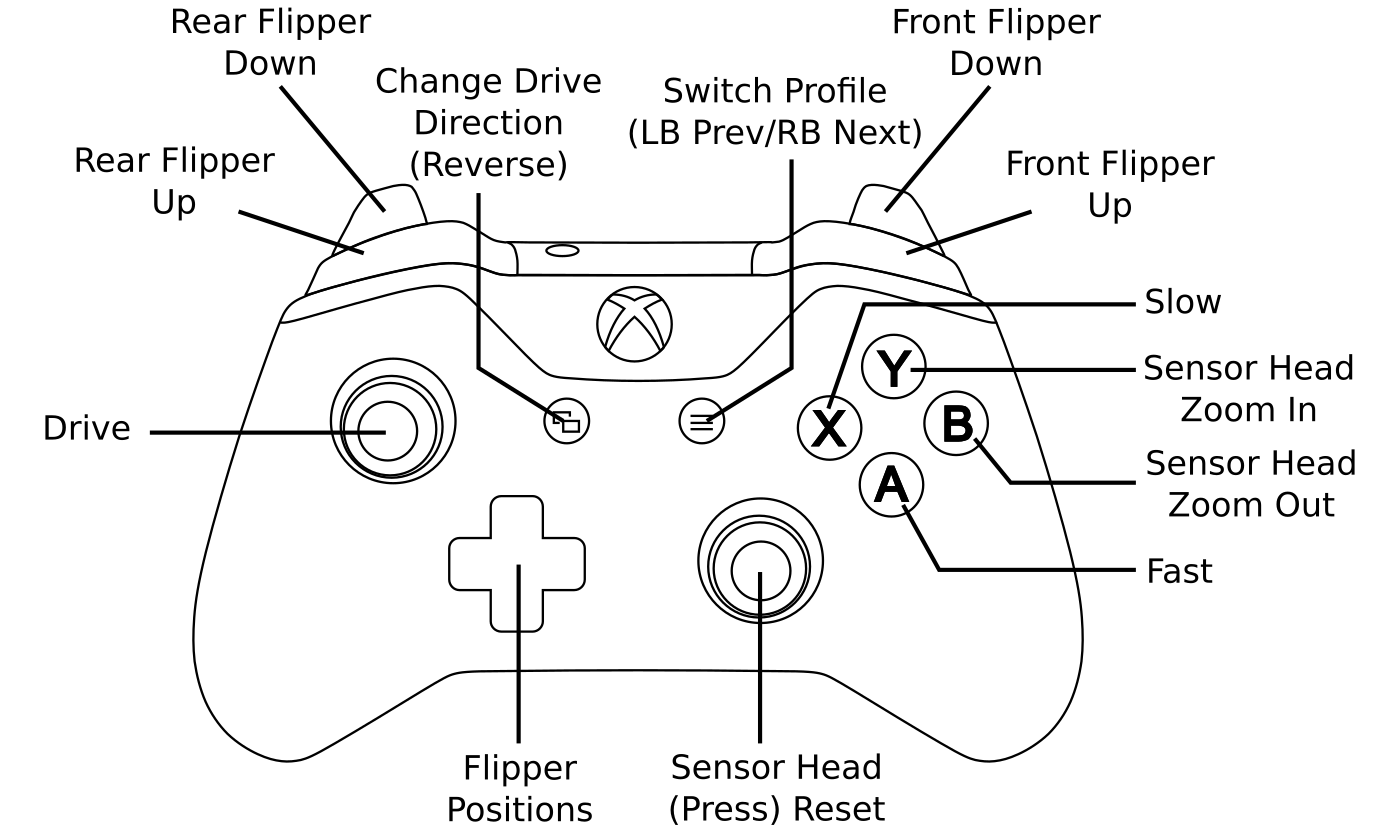}
  \hspace{0.01\linewidth}
  \includegraphics[width=0.4\linewidth]{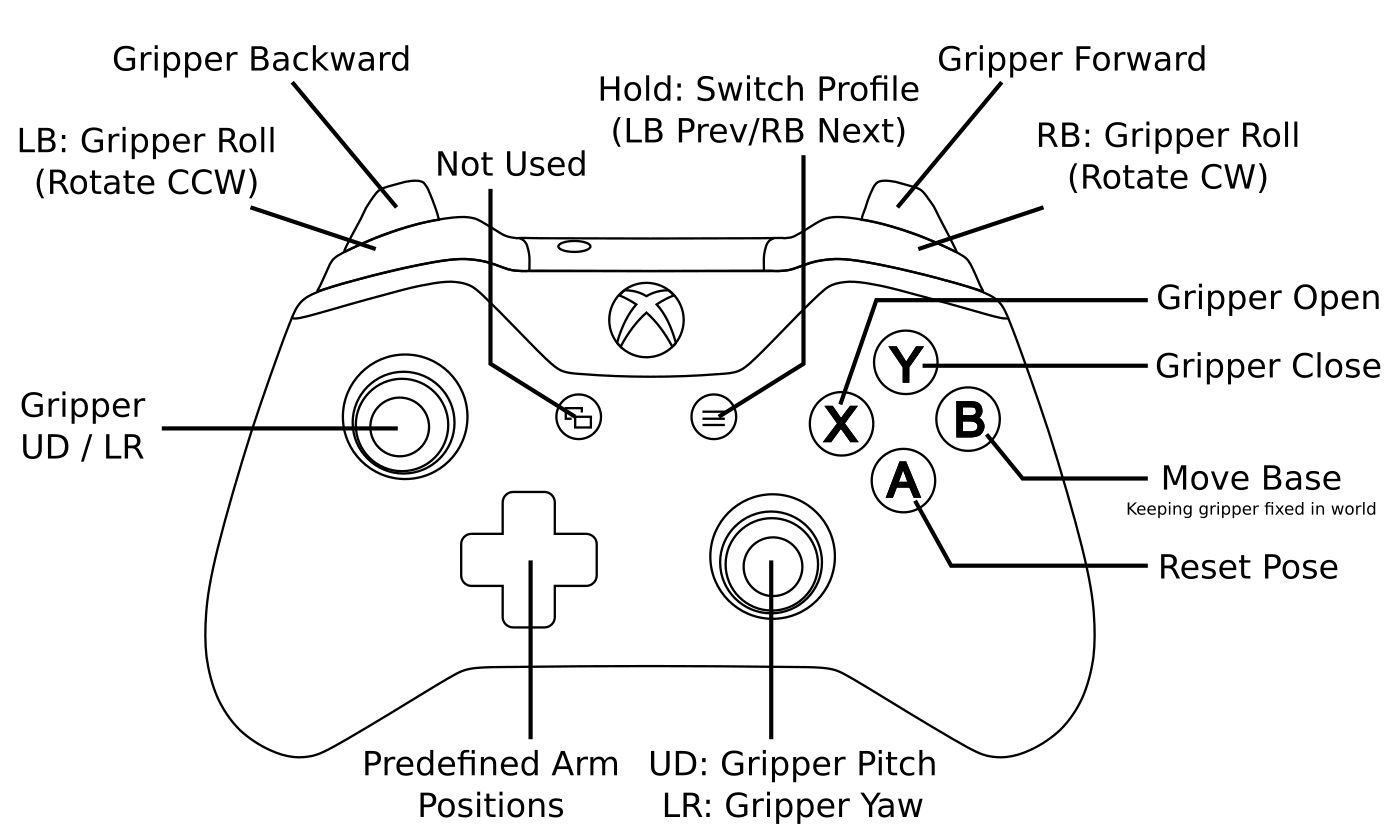}
  \end{center}
  \caption{The gamepad configurations for driving the robot (left) and using the gripper to inspect or manipulate objects (right).}
  \label{fig:gamepadcontrol}
\end{figure*}

\subsection{3D View}
Depending on the robot platform and mission environment, we use different \rviz~display plugins for data visualizations in the 3D view.
E.g., for our Asterix robot (see Fig.~\ref{fig:platforms}), we use a colored pointcloud \cite{oehler2021omnivision} (see Fig.~\ref{fig:asterix}) to allow for depth perception and semantic understanding.
As the LiDAR and camera resolution of the eC Scout robot (see Fig.~\ref{fig:platforms}) is too high to process for this use case, we use a voxel-downsampled cropped pointcloud for depth perception during navigation.

\subsubsection*{Tools}
For some actions, the robot requires input from the operator in the 3D view.
We use \rviz~Tool plugins to give this input.
To maximize the available screen space, we have built a tool control as an overlay on top of the 3D scene in the bottom left.

Next to the interact tool to interact with the 3D view, e.g., move the camera position and orientation, we have a tool to add 3D waypoints for the autonomous navigation function and robot actions that require 3D points as input, and a look-at-tool that defines a 3D point with a direction from which we want to inspect.

\subsubsection*{Sensor Values}
Sensor readings are displayed in the top left corner as shown in Fig. \ref{fig:scout_enrich} (1).
If multiple sensors are relevant to the mission, they are stacked horizontally.
The value range of a sensor value can have three classifications: safe (green), warning (orange), or dangerous (red).
Additionally to a color change, in the dangerous state, the sensor value will also increase slightly in size to draw the operator's attention without appearing too distracting.

\begin{figure}[t]
  \vspace{2mm}
  \begin{center}
    \begin{subfigure}{0.4\linewidth}
      \centering
      \includegraphics[width=\linewidth]{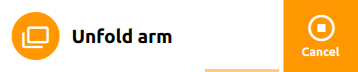}
      \caption{}
    \end{subfigure}
    \hspace{5mm}
    \begin{subfigure}{0.4\linewidth}
      \centering
      \includegraphics[width=\linewidth]{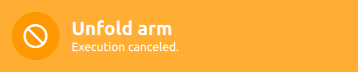}
      \caption{}
    \end{subfigure}
    \par\bigskip
    \begin{subfigure}{0.4\linewidth}
      \centering
      \includegraphics[width=\linewidth]{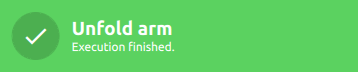}
      \caption{}
    \end{subfigure}
    \hspace{5mm}
    \begin{subfigure}{0.4\linewidth}
      \centering
      \includegraphics[width=\linewidth]{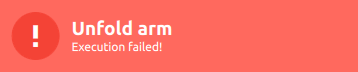}
      \caption{}
    \end{subfigure}
  \end{center}
  \caption{The action status during execution (a) and after term\-ination with status canceled (b), successful (c) or failed (d).}
  \label{fig:statuscolors}
\end{figure}

\subsubsection*{Active Task}
In Fig. \ref{fig:scout_enrich} (2) the active task of the robot is displayed.
The task progress indicator, as well as other important notifications, appears by moving into view from the top to draw attention by abruptly appearing in conjunction with onset movement~\cite{yantis1984abruptonset,abrams2003motiononset,rea2017movers} when a functionality execution, e.g. an autonomous behavior, starts.
During execution, it will only change the optional status text and a small moving progress bar to not distract the operator further.
When the execution finishes, to draw the operator's attention and quickly indicate the status, the color is changed as shown in Fig. \ref{fig:statuscolors} to one of the following outcomes: failed (red), canceled (orange), successful (green).

\subsubsection*{E-Stop Status}
The status of the emergency stops is displayed at the bottom center of the 3D view (see Fig. \ref{fig:asterix} (6)) if at least one e-stop is currently pressed.
The element can be expanded to show the status of each e-stop.
We use the package \textit{e\_stop\_manager}\footnote{\url{https://github.com/tu-darmstadt-ros-pkg/e_stop_manager}} to manage the state of multiple e-stops.
Note that hardware e-stops report their state to the e-stop manager but trigger independently to avoid dependence or a delay in critical situations.

%% file: sections/control.tex
The robot can be controlled in three control levels: \textit{Supervisory}, \textit{Direct}, and \textit{High-Level}.
In this section, the three control levels and our implementation is detailed.

\subsection{Supervisory Control}
The robot performs a mission autonomously and the operator is only serving in a supervisory role, watching and interpreting the available fused sensor data to gather information or take over control if the robot deviates from expected behavior.

The operator is informed about the planned navigation path, can see the task list with status, and has the option to pause or stop the mission execution at any time.
\new{Interaction with the 3D view and cameras is not limited in any way.
The autonomous behavior may request operator interaction, e.g., confirmation of a potentially detected victim.
However, to modify the robot's behavior in ways not directly supported by the current autonomous behavior or change the mission, the execution has to be halted.}

\subsection{Direct Control}
\label{sec:gamepadcontrol}
The operator can directly control the robot using a gamepad.
Minimal assistance functions such as inverse kinematics for manipulation are available.
We use separate configurations for driving and manipulation as shown in Fig.~\ref{fig:gamepadcontrol}.
The driving configuration includes a reverse mode which inverts the forward and backward direction.
In the user interface, the activation of this mode is indicated with an \textbf{R} in the controller mode visualization.

\subsection{High-Level Control}
\label{sec:highlevelcontrol}
The operator commands the robot to perform basic actions autonomously -- e.g., moving to a waypoint or looking at a point of interest -- but keeps high-level control over the robot's behavior.
This control mode is often used in combination with direct control.
For example, a basic action such as \textit{Look at} -- which computes and moves the end-effector such that it looks at a given 3D point from a given direction -- could be used to bring the manipulator in a good starting position to manually perform a manipulation task.

%% file: sections/configuration.tex
The UI can be used for different types of robot platforms with different capabilities.
Currently, all of the robots for which it has been applied (see Fig.~\ref{fig:platforms} for the ones used in this paper) are ground robots with either tracked, wheeled, or legged locomotion and also different in the available sensors and capabilities, e.g., only some of the robots have a manipulator arm.
In this section, we explain the configuration options to adapt to different platforms and changing sensors and functionalities.
We will give a short introduction to the used framework before covering how the interface can be adapted to the different platforms.

\begin{figure}
  \vspace{2mm}
  \includegraphics[width=\linewidth]{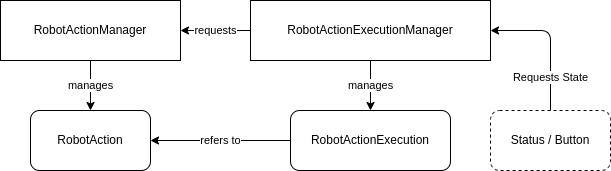}
  \caption{The class structure of singletons (rectangular) and objects (rounded) managing the available robot actions and their execution.}
  \label{fig:actionstructure}
\end{figure}

\subsection{Framework}
As mentioned above, the package \hectorrvizoverlay~is used to render our user interface on top of the robotics visualization tool \rviz.
The library comes with support for QtWidget or QML-based interfaces.
We have recently switched to QML as it comes with hardware-acceleration and does not require compilation resulting in faster prototyping and development times.
Our library also supports optional live-reloading for quick development and prototyping by automatically reloading the interface when one of the source files changes without having to restart the application.
In contrast to QtWidget, QML also supports inline logic in the interface files, and using our open-source package \qmlrosplugin\footnote{ROS: \url{https://github.com/StefanFabian/qml_ros_plugin}\\ROS2:~\url{https://github.com/StefanFabian/qml_ros2_plugin}}, it is possible to directly add the logic to interact with a ROS system to the declaration of a button.

Controls are encapsulated in separate files and are configured using exposed properties that can be set on each instance.
These properties can range from simple types such as numbers or strings to complex JavaScript or Qt objects.
JavaScript objects are reactive in Qt which means that changing a property of a JavaScript object will propagate to all elements and properties that use this property.
This enables us to change the color scheme of the entire user interface by changing one variable.

It is not always possible to inject a required property down the element tree, e.g., to inform an action button that the action it represents is already running.
In this case, we use singletons that manage the state in one place.
As shown in Fig. \ref{fig:actionstructure}, we use a singleton that manages the available actions to ensure that all elements refer to the same robot action and a separate singleton that manages the execution of said actions.
Elements such as the \textit{Active Task} or an \textit{Action Button} query the singleton to obtain the execution state and listen for new executions.
When the state changes, the elements can reflect this by updating the status or in the case of the button, changing the state to a \textit{Cancel} button.

Some common and robot-independent controls such as the Robot Action structure, the \rviz~tools toolbar, the multi-camera view, and many more are released as a separate open-source package in the context of this work\footnote{\url{https://github.com/tu-darmstadt-ros-pkg/hector_qml_controls}}.

\subsection{Configuration}
\label{sec:configuration}
\begin{figure}
  \vspace{2mm}
  \begin{center}
    \includegraphics[width=0.45\linewidth]{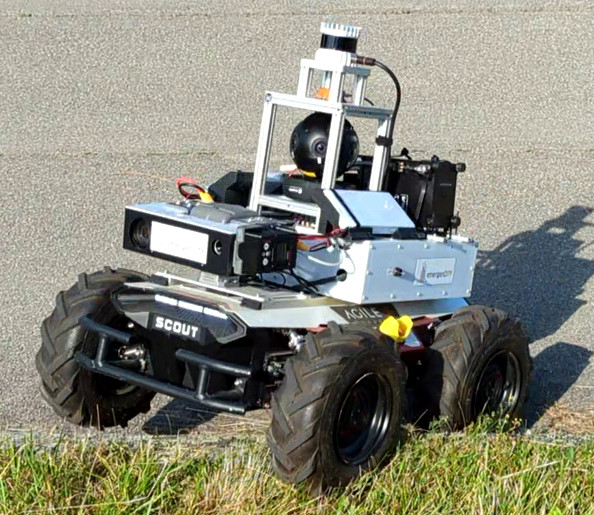}
    \includegraphics[width=0.475\linewidth]{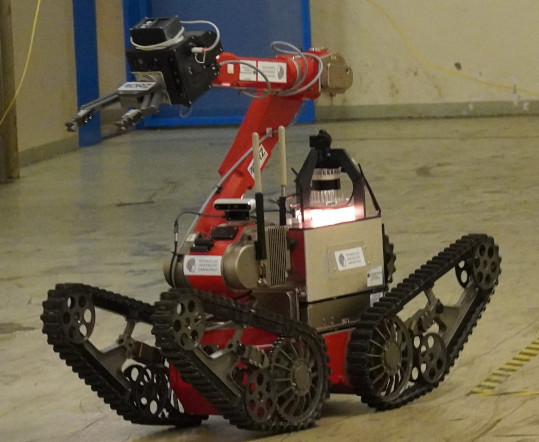}
    \includegraphics[width=0.27\linewidth]{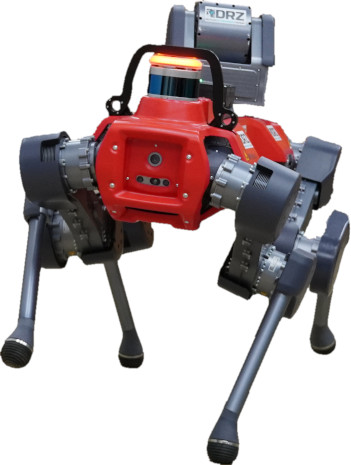}
    \includegraphics[width=0.27\linewidth]{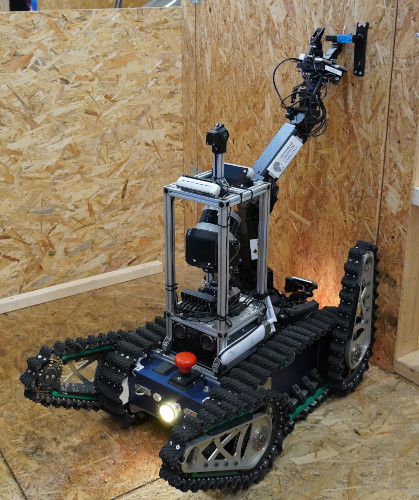}
    \includegraphics[width=0.27\linewidth]{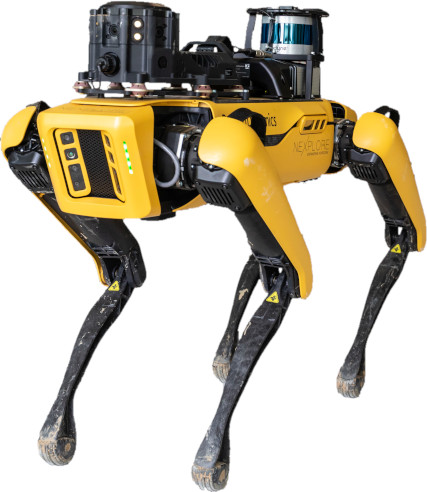}
  \end{center}
  \caption{The robot platforms the presented user interface is used with. From top to bottom, left to right: emergenCITY Scout, Telerob Telemax, ANYmal, Asterix, BD Spot.}
  \label{fig:platforms}
\end{figure}

The robots with which this interface is used and developed are shown in Fig.~\ref{fig:platforms}.
They each have different sensors, hardware, and capabilities.
Hence, to be able to adapt to the different robots and following \textbf{G.4.4} \textit{"Permit users tailor the display to their preferences"}, the interface is highly configurable.
General settings include the names of the flipper joints for tracked robots, the topics of robot state information such as battery and connectivity, etc.
Those are configured as display properties of the \rviz~display plugin that renders the interface and are only intended for developers.

\begin{figure}
  \vspace{2mm}
  \begin{center}
    \begin{subfigure}{0.45\linewidth}
      \includegraphics[width=\linewidth]{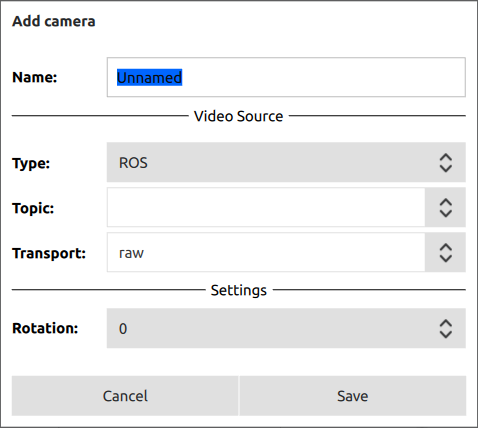}
      \caption{}
      \label{fig:cameradialog}
      \par\smallskip
      \includegraphics[width=\linewidth]{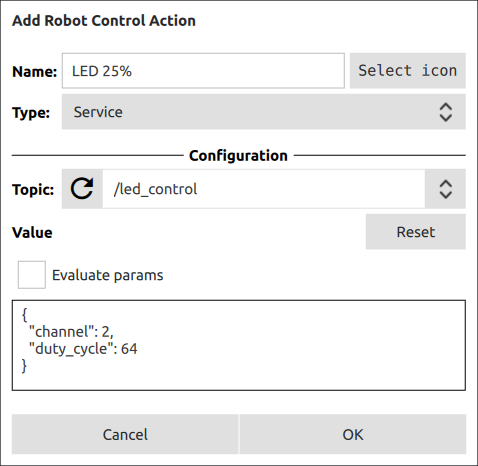}
      \caption{}
      \label{fig:addactiondialogservice}
    \end{subfigure}
    \begin{subfigure}{0.45\linewidth}
      \includegraphics[width=\linewidth]{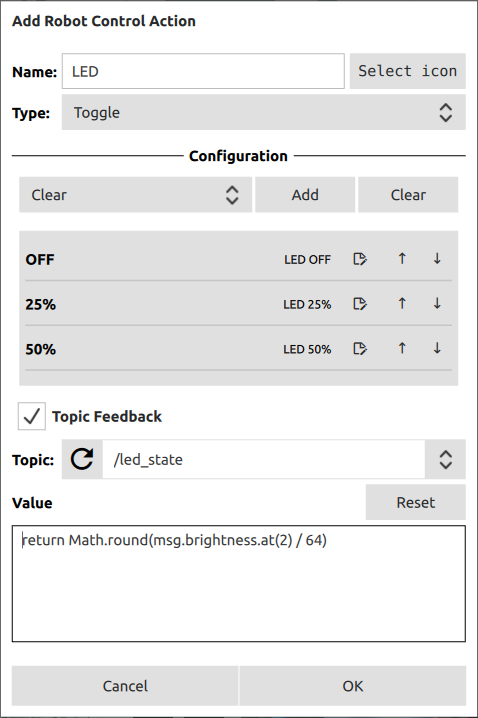}
      \caption{}
      \label{fig:addactiondialogtoggle}
      \vspace{1cm}
    \end{subfigure}
    \end{center}
  \caption{Examples of dialogs to add a camera (a), create an action that calls a ROS service (b), and create a composite toggle action with state parsed from a ROS topic (c).}
\end{figure}

The 3D view can be configured using the \rviz~plugin system.
Cameras can be added with a simple dialog as shown in Fig. \ref{fig:cameradialog}.
Available topics are identified using the \textit{sensor\_msgs/Image} topic type.
The robot actions can be added to the all actions list in the \textit{Behaviors} tab using the dialog shown in Fig. \ref{fig:addactiondialogtoggle}.
An action can be a ROS Communication, JavaScript code, a composite running multiple actions in sequence or parallel, or a toggle cycling through multiple actions.
Depending on the type additional configuration can be performed.
For a ROS action the request or message that is sent can be given either as JSON or a JavaScript function body can be specified that generates the message content dynamically, e.g., using data from an rviz tool to pass on as input.
A toggle action can have a feedback topic and a parsing function that extracts the current state from the received message thereby ensuring that a toggle such as LED 0\% / 50\% / 100\% always has the correct state in all open instances of the interface.

The configuration is saved in the \rviz~config file as a sub-property-tree of the UI display plugin.

%% file: sections/applications.tex
The UI presented in this work has been developed and improved based on experimental findings obtained in various robotics competitions. In these, the mission environment and tasks were beyond the authors' control and, in some cases, were previously unknown.
These competitions belong to two areas of application: \textit{inspection} and \textit{rescue}.
Both applications have many common requirements and can often be adapted to the specific use case with little or manageable effort.
In the context of inspection, we participated in and won the ARGOS Challenge in 2017 in France, as well as the plant disaster inspection challenge at the World Robot Summit 2018 in Japan.
In the context of rescue, we applied the UI with great success from 2017 to 2023 in the RoboCup Rescue Robot League competitions at the annual RoboCup German Open and the annual worldwide RoboCup event, as well as the bi-annual ENRICH Hackathon from 2017 to 2023.

\new{During competitions, the interface was used on laptops or monitors with screen sizes ranging from 14 to 27 inches, utilizing mouse and keyboard inputs.}

\begin{figure}
  \vspace{2mm}
  \includegraphics[width=\linewidth]{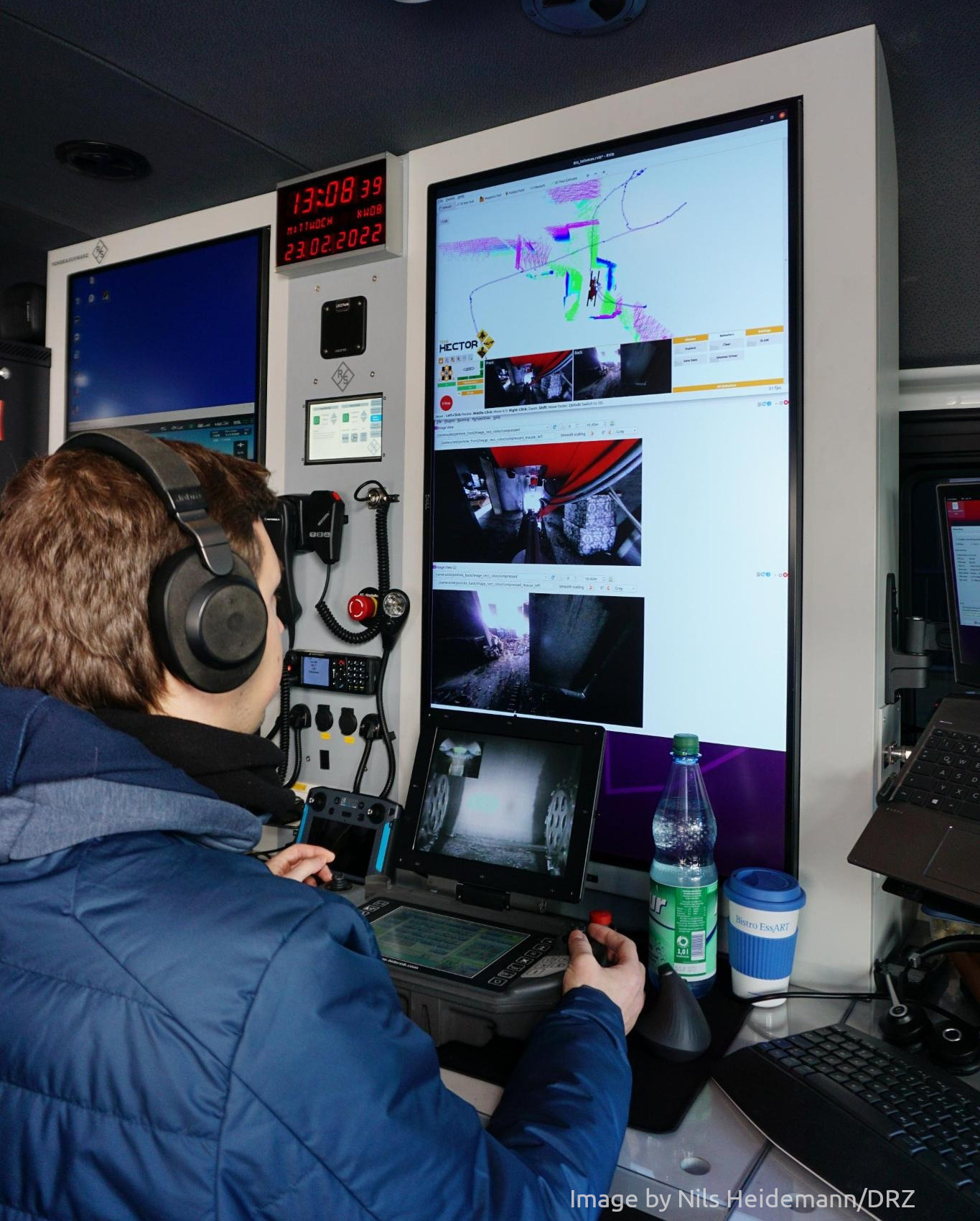}
  \caption{\new{The presented interface used in combination with a proprietary interface during a field deployment after a residential complex fire in Essen, Germany, 2022.}}
\end{figure}

\new{Outside of competitions, the interface was also used in a field deployment by the German Rescue Robotics Center Task Force where the lessons learned are summarized in \cite{surmann22}.}

In the following section, we will discuss the competitions in both contexts, the competition-specific adaptations where necessary, and summarize the  lessons learned during the competitions.

\subsection{Inspection}
\textbf{The ARGOS Challenge} was a competition sponsored by Total Energies and co-organized by the French research funding institution Agence Nationale de la Recherche (ANR) from 2013 to 2017 to develop the first autonomous inspection robot for the oil and gas industry.

The challenge was a major first driver for the development of a predecessor of the presented interface as it was a competition requirement that the user interface should be operable by end-users.
Two separate end-user interfaces were developed, one for the planning of inspection missions and a supervision interface to monitor and control the robot while executing the planned mission autonomously or, if needed, with operator interaction.
As both UIs were developed as \rviz~overlays, the implementation detail that they were separate interfaces could be hidden from the user and it was possible to seamlessly switch between the interfaces.

\textbf{The World Robot Summit} (WRS) was a competition hosted by the Ministry of Economy, Trade and Industry of Japan, and the New Energy Industrial Technology Development Organization in Tokyo, Japan 2018. The objective was to perform various pre-defined inspection missions on a test plant.

\begin{figure}
  \vspace{2mm}
  \begin{subfigure}{0.48\linewidth}
    \includegraphics[width=\linewidth]{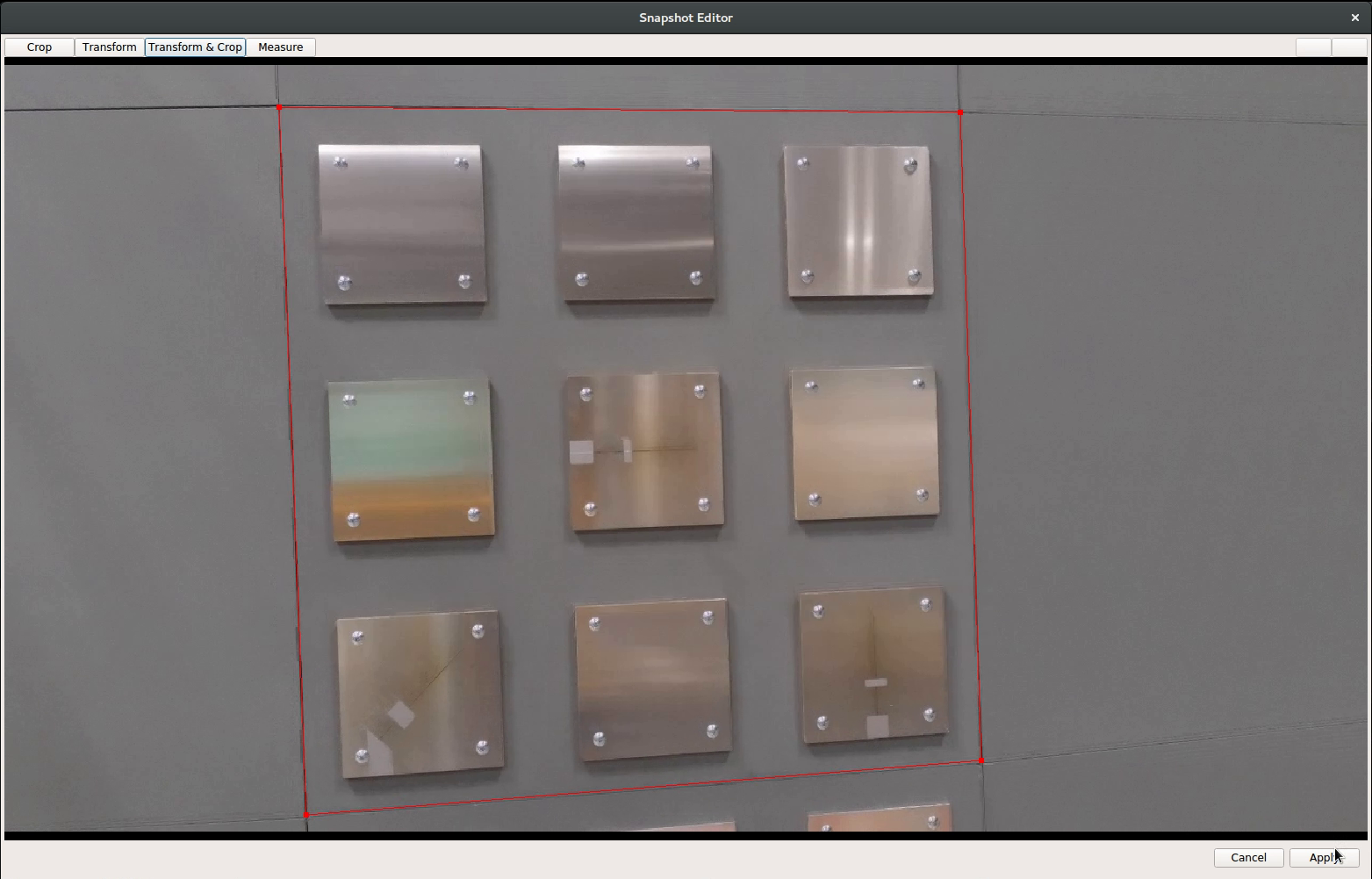}
  \end{subfigure}
  \begin{subfigure}{0.48\linewidth}
    \includegraphics[width=\linewidth]{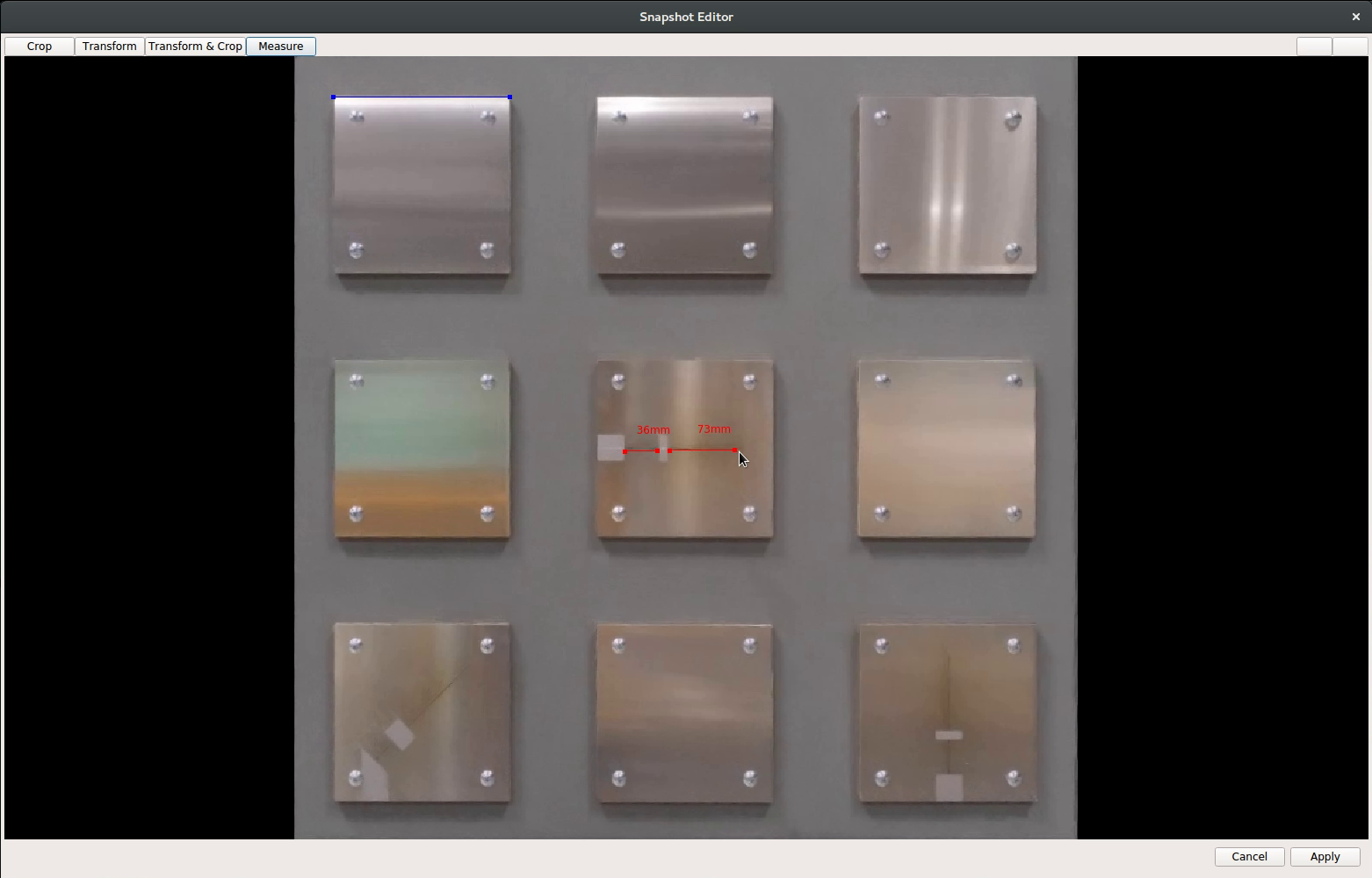}
  \end{subfigure}
  \caption{The snapshot editor to measure crack lengths in camera snapshots at the WRS competition.}
  \label{fig:snapshoteditor}
\end{figure}

In one of the missions, it was asked to measure the length of cracks on stone panels with known dimensions.
To make this user-friendly for the operator, we added a tool with low effort during the competition week that takes a snapshot via a button in the camera view and opens it in a separate window.
There the image can be transformed by clicking on the four corners of the panel to rectify the image taken from any angle.
Then, a known length was marked in the image to compute the pixel-to-\SI{}{\centi\metre} ratio and allow subsequent measurements by marking the start and end of each crack.
This process is shown in Fig.~\ref{fig:snapshoteditor}

\subsection{Rescue}
\textbf{The RoboCup Rescue Robot League} competitions at the RoboCup German Open and worldwide events are annual competitions where multiple standardized test arenas assess the performance of rescue robots with a mixture of mobility and dexterity tasks to evaluate the robot's capabilities in traversing difficult terrain and perform manipulation tasks, e.g., pressing a button or turning a lever~\cite{pellenz2014robocup}.

Points are scored by repeatedly completing the test arena in a limited time window of 20 minutes.
Since 2023, the mission time has been split into a task time of 10 minutes where the robot has to move from one end of the arena to the other overcoming obstacles, mapping, or opening a door, and another dexterity time of 10 minutes where the robot has to perform multiple different manipulation tasks, e.g., inspecting pipes and pressing buttons.

The time limit creates an artificially high stress level on the operator and creates a good environment to test the effectiveness of operator assistance functions in reducing operator stress.
It should be noted that the type of stress and mental load is also relevant for testing assistance functions, as in some rescue missions the time may be less critical than ensuring the proper operation of the robot and avoiding mission failure due to tip-over or hardware failure resulting in a possible loss of the platform and mission failure.
In the RoboCup Rescue Robot League hardware failure or tip-over events reduce the effective mission time and resets to repair the robot come with a time penalty for the current mission but the robot is not lost and repeating a mission is possible.
This leads to a higher inclination for high-risk maneuvers and less attention paid to assistance functions that would reduce tip-over risk at the cost of time.

\textbf{The ENRICH robotics hackathon} is a bi-annual competition where the objective is to perform different missions, like exploration and mapping, in a nuclear power plant with multiple radioactive probes that have to be located and mapped.

The environment is less cluttered than the RoboCup test arenas and mobility challenges are limited to typical human environments, e.g., stairs, floor openings, or narrow doorways.
Challenges are introduced by varying lighting conditions from well-lit rooms to dark corridors and limited wireless infrastructure in certain areas requiring autonomous navigation to avoid getting stuck.

We participated with two robot platforms: the wheeled eC Scout robot for fast exploration of areas with flat ground and the Telemax robot with stair climbing ability for manipulation and precise radiation source localization.
A single operator and UI were used per robot.